\documentclass[conference]{IEEEtran}
\usepackage[utf8]{inputenc}
\usepackage{graphicx}
\usepackage{amsmath, amssymb}
\usepackage{multirow}
\usepackage{hyperref}
\usepackage{booktabs}
\usepackage{tabularx}

\usepackage[numbers]{natbib}

\title{CrunchLLM: Multitask LLMs for Structured Business Reasoning and Outcome Prediction}

\author{
    \IEEEauthorblockN{Rabeya Tus Sadia}
    \IEEEauthorblockA{
        Department of Computer Science \\
        University of Kentucky \\
        rabeya.sadia@uky.edu
    }
    \and
    \IEEEauthorblockN{Qiang Cheng\footnotemark *}
    \IEEEauthorblockA{
        Department of Computer Science, \\
        Institute for Biomedical Informatics \\
        University of Kentucky \\
        qiang.cheng@uky.edu
    }
}
\begin{document}
\maketitle

\begin{abstract}
Predicting the success of start--up companies, defined as achieving an exit through acquisition or IPO, is a critical problem in entrepreneurship and innovation research. 
Datasets such as Crunchbase provide both structured information (e.g., funding rounds, industries, investor networks) and unstructured text (e.g., company descriptions), but effectively leveraging this heterogeneous data for prediction remains challenging. 
Traditional machine learning approaches often rely only on structured features and achieve moderate accuracy, while large language models (LLMs) offer rich reasoning abilities but struggle to adapt directly to domain-specific business data.  
We present \textbf{CrunchLLM}, a domain-adapted LLM framework for startup success prediction. CrunchLLM integrates structured company attributes with unstructured textual narratives and applies parameter-efficient fine-tuning strategies alongside prompt optimization to specialize foundation models for entrepreneurship data. 
Our approach achieves accuracy exceeding 80\% on Crunchbase startup success prediction, significantly outperforming traditional classifiers and baseline LLMs. 
Beyond predictive performance, CrunchLLM provides interpretable reasoning traces that justify its predictions, enhancing transparency and trustworthiness for financial and policy decision makers.  This work demonstrates how adapting LLMs with domain-aware fine-tuning and structured--unstructured data fusion can advance predictive modeling of entrepreneurial outcomes. CrunchLLM contributes a methodological framework and a practical tool for data-driven decision making in venture capital and innovation policy.
\end{abstract}

\section{Introduction}
Predicting the success of startup companies is a fundamental problem in entrepreneurship and innovation research. 
Investors, policymakers, and entrepreneurs seek reliable methods to identify which startups are most likely to succeed, typically measured by major exits such as acquisitions or IPOs. 
Datasets such as Crunchbase provide rich information on startups, including structured variables like funding history and industry category, as well as unstructured data such as company descriptions \cite{crunchbase}. 
However, building predictive models that can effectively integrate such heterogeneous information remains a significant challenge.  

Traditional machine learning approaches have focused on structured features derived from Crunchbase, applying models such as logistic regression, random forests, or shallow neural networks to predict outcomes \cite{zbikowski2021machine, kim2023succeed, thirupathi2021machine, razaghzadeh2024predicting}. 
While these approaches achieve moderate accuracy, they are limited in their ability to capture the nuanced signals embedded in textual company descriptions or investor narratives. 
At the same time, recent interest in using large language models (LLMs) for business analytics has highlighted both opportunities and obstacles. 
General-purpose LLMs such as GPT-3, PaLM, and LLaMA have demonstrated strong reasoning and text understanding capabilities \cite{Brown2020GPT3,Chowdhery2022PaLM,touvron2023llama}, but when directly applied to Crunchbase data, their predictive accuracy is often only marginally better than random guessing \cite{bornstein2023prediction}. 
The challenges include (i) effectively adapting models trained on broad web corpora to specialized business domains, (ii) integrating structured and unstructured data representations, and (iii) addressing issues of limited labeled data for training domain-specific predictors.  

These challenges motivate our development of \textbf{CrunchLLM}, a domain-adapted LLM framework for startup success prediction. 
Our key innovation lies in fine-tuning foundation LLMs using parameter-efficient strategies \cite{Hu2022LoRA,dettmers2023qlora} tailored to Crunchbase data. 
We design a pipeline that jointly leverages structured company attributes and unstructured natural language descriptions, enabling the model to capture both quantitative business signals and qualitative narratives. 
In addition, we explore prompt optimization and task-specific fine-tuning to adapt general-purpose reasoning abilities of LLMs to entrepreneurship prediction tasks. 
This approach not only improves predictive accuracy, achieving performance exceeding 80\% on Crunchbase success prediction, but also provides interpretable reasoning traces that justify model outputs, an important requirement in financial and policy applications.  

By bridging advances in large language models with the practical needs of entrepreneurship research, CrunchLLM contributes both a methodological framework and an applied tool. 
It demonstrates how domain adaptation, parameter-efficient fine-tuning, and structured--unstructured data integration can be combined to yield significant gains in startup success prediction, offering new opportunities for data-driven decision making in venture capital and innovation policy.  

In brief, our main contributions and novelties include the following:
\begin{itemize}
\item We propose CrunchLLM, an adaptive LLM-based framework for predicting startup success from Crunchbase, integrating structured attributes with unstructured narratives.
\item We introduce prompt-guided multimodal fusion, a multi-task approach that dynamically combines structured tabular signals with company-profile text, enabling the model to identify prediction-relevant evidence from large-scale corpora, addressing challenges in existing LLMs that struggle with extracting key contents across modalities.
\item Our domain adaptation pipeline represents a novel strategy that integrates parameter-efficient fine-tuning and prompt optimization, tailored to entrepreneurship and innovation tasks where general-purpose LLMs typically underperform.
\item We demonstrate that CrunchLLM achieves state-of-the-art results on Crunchbase benchmarks, showing statistically significant gains over tabular-only, text-only, and generic LLM baselines (paired t-tests, $p<0.01$), while producing human-readable reasoning traces that enhance interpretability for investment, research, and policy-making.
\end{itemize}

\section{Related Work}

\subsection{Large Language Models}
Large Language Models (LLMs) have transformed artificial intelligence research by demonstrating that scaling parameters and training data enables emergent reasoning abilities and strong performance on diverse downstream tasks \cite{Kaplan2020Scaling,Wei2022Emergent}. Models such as GPT-3, PaLM, and GPT-4 have been applied across domains ranging from natural language understanding to reasoning and decision support \cite{Brown2020GPT3,Chowdhery2022PaLM}. Recent open-source models, including LLaMA and DeepSeek, have further extended access to foundation models with competitive reasoning capabilities \cite{touvron2023llama,deepseek2025}. These developments motivate adapting LLMs for domain-specific prediction tasks such as startup success forecasting, where data are largely textual and semi-structured.  

\subsection{Fine-Tuning for Domain Adaptation}
Fine-tuning is widely used to adapt LLMs trained on general corpora to specialized tasks. Full fine-tuning can be computationally expensive, leading to increasing interest in parameter-efficient fine-tuning methods such as Low-Rank Adaptation (LoRA) \cite{Hu2022LoRA} and QLoRA \cite{dettmers2023qlora}. These methods reduce the number of trainable parameters while preserving model performance, making them attractive for adapting LLMs to structured business data. Applications of LoRA \cite{Hu2022LoRA} and related strategies have shown effectiveness in specialized domains such as medicine, finance, and scientific reasoning \cite{med42,curie2025}, demonstrating their utility for contexts where domain-specific expertise is critical. Multi-stage fine-tuning approaches, where models are progressively trained from basic domain concepts to task-specific reasoning, have also been shown to improve factuality and robustness \cite{Hu2025Legal,curie2025}. Inspired by these findings, CrunchLLM leverages parameter-efficient fine-tuning and domain adaptation to specialize LLMs for entrepreneurship and innovation data.  

\subsection{Predictive Modeling of Startup Success}
The prediction of startup outcomes has been studied through traditional machine learning approaches using structured datasets such as Crunchbase. Prior work has employed classifiers including logistic regression, random forests, and neural networks to predict company exits, IPOs, or acquisitions, often relying on handcrafted features such as funding history, team composition, and market indicators \cite{thirupathi2021machine,zbikowski2021machine, kim2023succeed}. While these methods provide some predictive capability, they are limited in incorporating the rich unstructured text data (e.g., company descriptions, press releases) that often signal future performance. Recent research has begun exploring transformer-based architectures for business prediction tasks, but challenges remain in effectively integrating structured and textual data at scale, and in particular, reliable prediction of startup companies' success is lacking. CrunchLLM addresses this gap by fine-tuning LLMs on Crunchbase data to capture both structured business signals and unstructured textual narratives, thereby improving prediction accuracy while also leveraging the reasoning abilities of LLMs to provide interpretable justifications for predictions.

\section{Methodology}
\subsection{Dataset}
The dataset used in this study was obtained from the Crunchbase database\cite{crunchbase}, a widely recognized platform that provides business information on private and public companies, including details on founders, executives, investors, and funding activities. For this research, we applied for and were granted access to the Crunchbase database. The version of the data used in our experiments was retrieved on June 11, 2025. In the following section, we outline the preprocessing steps and dataset preparation procedures, and the workflow pipeline employed in our study.
We preprocess the Crunchbase \cite{crunchbase} dataset into structured chat prompts for binary classification (success vs. non-success) and justification generation. Balanced and cleaned samples are prepared to prevent leakage and ensure uniform evaluation.
\section{Dataset Preprocessing}

\subsection{Feature Engineering}

To facilitate effective language model-based reasoning over structured startup profiles, we constructed a cleaned and task-aligned dataset from the Crunchbase database. Crunchbase is a widely-used platform in entrepreneurship research, offering extensive firm-level information including investment history, personnel, and major company events \cite{crunchbase}.

We leverage multiple relational tables from the Crunchbase database to construct a comprehensive company-level dataset. These include metadata such as unique identifiers, company names, descriptions, and founding dates; records of liquidity events including Initial Public Offerings (IPOs) and acquisitions; detailed information on funding rounds, including capital raised and investor participation; and employment data used to infer the number of executive-level personnel. This multi-source integration allows us to extract rich temporal, financial, and human capital features for downstream analysis.

To capture firm maturity, we engineered a temporal feature representing organizational age by calculating the time difference between each entity’s creation date and the current date. This feature serves as a proxy for company longevity, which has been shown to correlate with business outcomes such as growth potential and investment attractiveness \cite{piva2002determinants}.
Following recent studies \cite{bornstein2023prediction}, we define a company as ``successful'' if it has either undergone an IPO or been acquired. This binary definition is widely adopted in entrepreneurship and venture capital research, as such exits represent significant financial or strategic milestones. These events typically mark a company’s transition to maturity or integration into a larger entity, often associated with substantial value realization for founders and investors.

Ewens and Rhodes-Kropf \cite{ewens2015vc} use IPOs and acquisitions as outcome variables to study VC performance and innovation dynamics. Similarly, Gornall and Strebulaev \cite{gornall2020vc} treat these exits as standard success measures in the context of private firm valuation. In our study, we also followed the same strategy with the Crunchbase dataset.


We derived several high-level financial and organizational indicators to support structured reasoning tasks. These include total capital raised, the number of funding rounds a company has completed, the cumulative number of distinct investors involved, the count of acquisitions the company has made, and the estimated number of executive-level employees. These features capture both the financial trajectory and human capital profile of each organization, offering valuable context for downstream predictive modeling and analysis.

To ensure broad data coverage, we retained all organizations in the dataset regardless of missing values and applied robust imputation strategies to handle incomplete entries. Specifically, missing numeric features were imputed with zeros to maintain consistency, while organizational age was set to –1 when unavailable to indicate absence explicitly. Additionally, binary indicators were cast to integers and defaulted to 0 in cases of missing values, ensuring compatibility with model inputs without introducing bias.


\subsection{Data Challenges for Language Models}

\begin{figure}[h]
    \centering
    \includegraphics[width=0.45\textwidth]{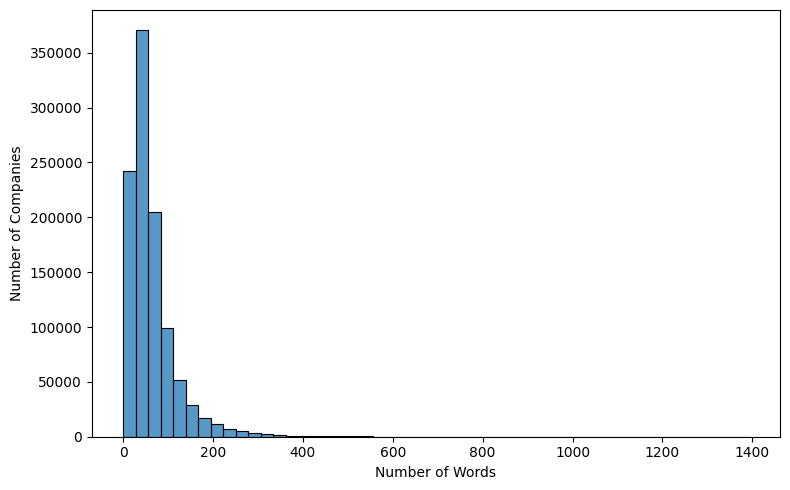}
    \caption{Distribution of description length (in tokens).}
    \label{fig:desc_length}
\end{figure}

To better understand the limitations of applying LLMs to structured business profiles, we examined the data from both linguistic and distributional perspectives. Figure~\ref{fig:desc_length} illustrates the distribution of description lengths (in tokens), highlighting a wide range of input lengths across entities. This token length heterogeneity poses challenges for LLMs, especially those with fixed context windows, as overly long or short inputs can lead to truncation or underutilization of model capacity.

\begin{figure*}[h]
    \centering
    \includegraphics[width=\textwidth]{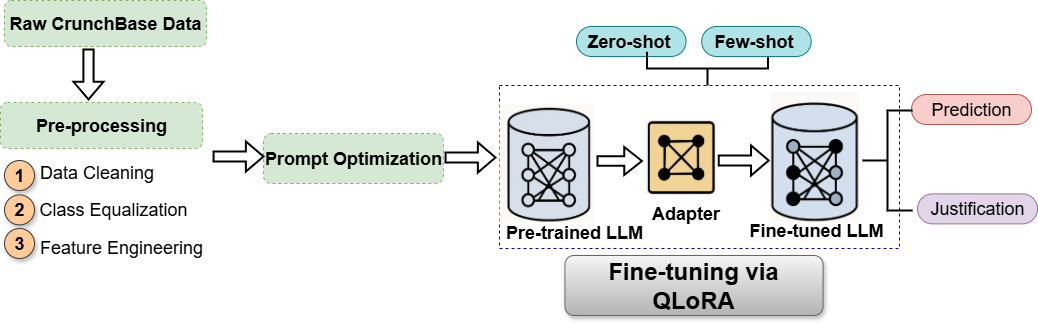}
    \caption{CrunchLLM overall Workflow.}
    \label{fig:workflow}
\end{figure*}

Additionally, the distribution of the success labels reveals a significant class imbalance between successful and unsuccessful entities. This imbalance can skew the learning process, leading to biased predictions and reduced confidence for minority-class examples. Addressing these issues is crucial for ensuring fairness and robustness in downstream LLM-based tasks such as classification or justification generation. The overall CrunchLLM workflow is summarized in Figure \ref{fig:workflow}, illustrating the data preprocessing, prompt construction, parameter-efficient finetuning, and evaluation pipeline.

\subsection{Multitask finetuning}

\subsubsection{Prompt Optimization for Supervised Fine-Tuning (SFT)}

We follow recent literature on instruction tuning \cite{alpaca, ouyang2022training, chung2022scaling} to construct prompt-response pairs in a chat format for multitask SFT. Each instance includes structured business metadata and a multi-step instruction to guide the LLM to perform both classification and justification. Our instruction tuning strategy includes the following key design choices:

\begin{itemize}
    \item \textbf{Chat-style Prompting}: Each example is formatted using the \texttt{<|im\_start|>} and \texttt{<|im\_end|>} special tokens, allowing alignment with the chat-based architecture of instruction-tuned LLMs such as Qwen, LLaMA-2-chat, GPT.
    
    \item \textbf{Multitask Instruction}
    The user instruction explicitly specifies two tasks: (i) predicting the binary success label, and (ii) providing a brief justification based on the provided profile. This aligns with multitask prompting methods.
    
    \item \textbf{Structured Input Encoding}
    Tabular metadata (e.g., age, funding, investors, executive count) is converted into a formatted profile-style text block, ensuring consistency across samples while preserving the original semantics of each feature.
    
    
    \item \textbf{Grounded Justification} For supervised fine-tuning, the justification is templated from label-aligned heuristics (e.g., strong funding and large executive team for successful firms). This provides stable learning targets for explanation generation.
    
    \item \textbf{Balanced Training Distribution} To ensure robust generalization, the dataset is balanced across success classes, mitigating the impact of class imbalance during training.
\end{itemize}

This design enables the model to learn structured reasoning over tabular inputs and generate natural language justifications in a controlled format. Such multitask tuning has shown to enhance both accuracy and interpretability in real-world decision-making settings \cite{selfinstruct, zhou2022least}.

\subsection{Zero-shot inference}
To evaluate model generalization without fine-tuning, we perform zero-shot inference using open instruction-tuned models. Each input prompt encodes structured business metadata along with a classification and justification instruction. 
\subsection{Few-shot Supervision}
We further investigate how few-shot fine-tuning improves performance. Here, we fine-tune each model on subsets of 1k, 2k, and 4k examples using multitask chat-style prompts. These prompts contain both label and justification targets.

\section{Implementation Details}

To ensure reproducibility and comparability across models, we describe here the complete fine-tuning setup used for CrunchLLM. 

\paragraph{Tokenizer and Input Formatting} 
All datasets were reformatted into a chat-style dialogue consistent with instruction-tuned LLMs. The \texttt{user} message contained the structured company profile, and the \texttt{assistant} message contained both the classification label and a short justification. We adopted the special tokens \texttt{<|im\_start|>} and \texttt{<|im\_end|>} to clearly delimit conversational turns. Each input was truncated or padded to a maximum length of 256 tokens. Since some models (such as GPT-2) do not provide a dedicated padding token, we reused the end-of-sequence (EOS) token as padding for consistency across model families.

\paragraph{Parameter-Efficient Fine-Tuning}
We applied QLoRA \cite{dettmers2023qlora}, which combines low-rank adapters with 4-bit quantization, to all model backbones. Quantization was performed using the NF4 scheme with \texttt{bfloat16} computation. LoRA adapters were inserted into different components depending on the model architecture: for Qwen and LLaMA, we targeted the \texttt{q\_proj} and \texttt{v\_proj} attention projections, while for GPT-2 we targeted the combined \texttt{c\_attn} module, which merges query, key, and value projections. In all cases, the LoRA rank was set to 16 with scaling factor $\alpha=16$ and dropout $=0.1$. This configuration balances parameter efficiency with sufficient adaptation capacity.
\paragraph{Prompt Optimization Principles}

\begin{figure*}[]
    \centering
    \includegraphics[width=0.6\textwidth]{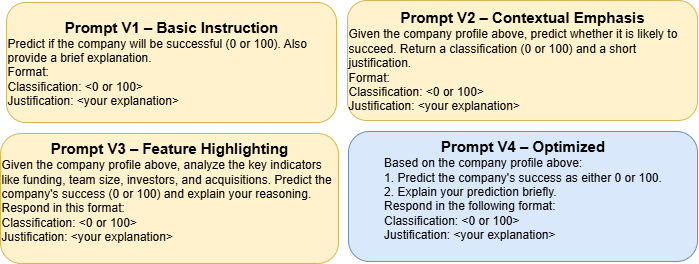}
    \caption{Prompt optimization ablation for classification and justification. Four prompt variants were explored from basic instruction (V1) to an optimized format (V4) to assess their impact on model performance. The optimized prompt (V4) showed improved alignment with expected output structure and interpretability.}
    \label{fig:prompt}
\end{figure*}

The refinement of prompts from V1 to V4, as shown in Figure \ref{fig:prompt}, was guided by
principled rules of instruction tuning rather than ad-hoc trial and error.
Following recent literature on systematic prompt engineering, we adopted
an iterative design philosophy where each modification was motivated by
clear principles and intermediate feedback signals. First, we applied the
\textit{clarity and specificity} rule: separating prediction and justification
instructions to reduce ambiguity, similar to a forward-looking prompt
selection strategies used in sequential optimization frameworks
\cite{zhou2025sequential}. Second, we enforced \textit{structural consistency}
by encoding tabular metadata in a profile-style format, aligning with
frameworks such as the Prompt Canvas that emphasize structured input
design \cite{schick2024promptcanvas}. Third, we introduced \textit{grounded justification} by explicitly linking explanations to observable features (e.g., funding raised, executive count). This refinement was motivated by feedback from earlier prompts, where the model often produced generic or unsubstantiated justifications. By enforcing grounding, we reduced hallucinations and improved interpretability, consistent with feedback-driven iterative prompting methods \cite{srivastava2024sipe}. Finally, our refinements
were informed by empirical findings on prompt iteration practices in
enterprise settings, which show that practitioners frequently adjust
prompts to improve format compliance and stability before assessing
final accuracy \cite{liu2024enterprise}. In this way, each transition
(V1$\rightarrow$V2, V2$\rightarrow$V3, V3$\rightarrow$V4) was justified not
only by downstream results but also by established principles of prompt
optimization.

\paragraph{ Training Schedule}
All models were trained under the same optimization settings for fairness. Training was performed for 5 epochs using the AdamW optimizer with a cosine learning rate scheduler. The initial learning rate was set to $5 \times 10^{-4}$ with 20 warmup steps. A weight decay of 0.01 was applied for regularization. Due to GPU memory constraints, we used a per-device batch size of 1 with gradient accumulation of 2, yielding an effective batch size of 2. Mixed precision training with \texttt{bf16} was enabled to further improve memory efficiency and throughput. All experiments were run on an NVIDIA A100 GPU with 40GB of memory.

\section{Results Analysis}

\subsection{Models Evaluation}

We intentionally cover a spectrum of open-weight LLMs from ~1B to 8B parameters to study the trade-off between accuracy, interpretability, and compute/memory cost. 
All selected models have mature ecosystem support and allow parameter-efficient fine-tuning (PEFT; e.g., LoRA/QLoRA~\cite{Hu2022LoRA, dettmers2023qlora}), enabling reproducible adaptation on modest GPUs.
By including both ``classic'' and recent families, we can benchmark gains due to modern pretraining, tokenization, and instruction tuning while keeping licensing appropriate for research use.

In summary, this set of LLMs is selected because of (i) coverage of model scales (tiny to mid); (ii) open weights and active ecosystems; 
(iii) compatibility with PEFT/quantization for cost-aware fine-tuning; 
(iv) diverse pretraining vintages to reveal where modern LLMs help most on Crunchbase;  and
(v) fair, reproducible comparisons across size and family.

\begin{itemize}
  \item \textbf{Qwen-2.5-3B} - Compact, recent 3B decoder-only model aimed at efficient fine-tuning; strong general text understanding with low VRAM footprint, suitable as a small but competitive baseline.
  \item \textbf{Llama-3-8B} - Meta’s 8B Llama~3 variant with robust general performance and mature tooling; a widely used reference point for 8B-class models~\cite{touvron2023llama}.
  \item \textbf{TinyLlama} - Ultra-compact (~1B) model from a pretraining effort; stresses the limits of parameter-efficient tuning and serves as a low-compute baseline for ablations.
  \item \textbf{Llama-3.2-3B} - Newer 3B compact variant optimized for efficiency and deployment; evaluates how far a small modern checkpoint can be pushed with PEFT on structured+text business data.
  \item \textbf{GPT-2} - Classic 1.5B decoder-only Transformer~\cite{radford2019language} used as a historical baseline to highlight gains from contemporary pretraining and adaptation methods.
\end{itemize}

\subsection{Metrics}
We evaluate classification performance using Precision, Recall, F1, Accuracy (ACC), and Area Under the Curve (AUC). 
For justification generation, we adopt BERTScore (Precision, Recall, and F1) to assess the semantic quality of model explanations. 
We systematically benchmark multiple baselines across three training regimes: zero-shot, few-shot, and full-data fine-tuning. 

The results in Table~\ref{tab:prediction_result} and Table~\ref{tab:justification_result} confirm that full-data fine-tuning consistently yields the best performance across both prediction and justification tasks. 
Among baseline models, LLaMA-3-8B and LLaMA-3.2-3B achieve strong classification results ($>$0.79 F1) with competitive justification quality. 
However, our proposed CrunchLLM substantially outperforms all baselines, surpassing  F1 and Accuracy in prediction, while also delivering the highest AUC score. 
Notably, CrunchLLM also produces the most precise justifications, achieving superior BERT-Precision (0.83), reflecting its ability to ground explanations in meaningful business attributes.

Interestingly, GPT-2, despite being the smallest LLM evaluated, still surpasses XGBoost, showing that pretraining on language corpora transfers effectively to structured prediction. 
XGBoost, while competitive in classification (0.72 ACC), cannot generate textual justifications, underscoring the advantage of generative models in multitask reasoning. 
Overall, there is a clear correlation between a model’s predictive accuracy and the quality of its justifications: stronger classifiers also yield more coherent and semantically faithful explanations.

Table~\ref{tab:zero_vs_fewshot} further shows that all models benefit substantially from even minimal supervised exposure. 
Transitioning from zero-shot to 1k few-shot examples improves F1 and Accuracy by over 10\% in most cases, with CrunchLLM demonstrating the largest gains. 
These improvements highlight the value of domain-specific supervision and our model’s ability to exploit it efficiently. 
We also observe that F1 increases more sharply than Accuracy at early stages, likely due to class imbalance in few-shot subsets that amplifies recall for minority classes.

Overall, the results show that scaling model size consistently improves both classification accuracy and justification quality, while prompt-guided multitask fine-tuning is essential for aligning predictions with coherent explanations. Our domain-adapted CrunchLLM establishes state-of-the-art performance by jointly leveraging structured company data and profile-style text, achieving both high predictive accuracy and interpretable reasoning. This dual capability demonstrates the novelty of our approach, as it provides a unified framework that advances beyond existing machine learning baselines and generic LLMs by delivering reliable predictions together with transparent, contextually grounded justifications for entrepreneurship, investment, and policy applications.

\begin{table*}[h]
\centering
\caption{Prediction Performance Summary}
\label{tab:prediction_result}
\begin{tabular}{|l|c|c|c|c|c|}
\hline
Model & Precision & Recall & F1 & ACC & AUC \\
\hline
XGBoost        & 0.70 & 0.72 & 0.71 & 0.72 & 0.75 \\
GPT-2          & 0.73 & 0.75 & 0.74 & 0.72 & 0.78 \\
LLaMA-3-8B     & 0.80 & 0.82 & 0.81 & 0.82 & 0.84 \\
LLaMA-3.2-3B   & 0.78 & 0.80 & 0.79 & 0.80 & 0.84 \\
TinyLLaMA      & 0.75 & 0.77 & 0.76 & 0.78 & 0.81 \\
Qwen-2.5-3B    & 0.76 & 0.78 & 0.74 & 0.77 & 0.82 \\
\textbf{Our model} & \textbf{0.83} & \textbf{0.82} & \textbf{0.90} & \textbf{0.89} & \textbf{0.85} \\
\hline
\end{tabular}
\end{table*}

\begin{table*}[h]
\centering
\caption{Justification Quality Summary}
\label{tab:justification_result}
\begin{tabular}{|l|c|c|c|}
\hline
Model & BERT-Recall & BERT-Precision & BERT-F1 \\
\hline
GPT-2        & 0.82 & 0.76 & 0.78 \\
LLaMA-3-8B   & 0.82 & 0.76 & 0.78 \\
LLaMA-3.2-3B & 0.82 & 0.76 & 0.79 \\
TinyLLaMA    & 0.81 & 0.75 & 0.78 \\
Qwen-2.5-3B  & 0.81 & 0.79 & 0.80 \\
\textbf{Our model} & \textbf{0.80} & \textbf{0.83} & \textbf{0.81} \\
\hline
\end{tabular}
\end{table*}

\begin{table*}[h]
\centering
\caption{Comparison between Zero-shot and Few-shot Performance}
\label{tab:zero_vs_fewshot}
\begin{tabular}{|l|cc|cc|cc|cc|}
\hline
\multirow{2}{*}{Model} & \multicolumn{2}{c|}{Zero-shot} & \multicolumn{2}{c|}{Few-shot (1k)} & \multicolumn{2}{c|}{Few-shot (2k)} & \multicolumn{2}{c|}{Few-shot (4k)} \\
\cline{2-9}
 & F1 & ACC & F1 & ACC & F1 & ACC & F1 & ACC \\
\hline

LLaMA-3-8B    & 0.49  & 0.51 & 0.60 & 0.57 &  0.65  & 0.61     &  0.59  & 0.68         \\
LLaMA-3.2-3B   & 0.49 &  0.50  & 0.56 & 0.52  &  0.60  &  0.55   & 0.58   & 0.61  \\
TinyLLaMA     & 0.50  &  0.45 & 0.54  & 0.48  & 0.50    &  0.51     & 0.56     &   0.58      \\
Qwen-2.5-3B     & 0.46  & 0.47  & 0.52 & 0.54 &   0.52   & 0.56    & 0.54    & 0.60        \\
\textbf{Our model}    & 0.46  & 0.47 & \textbf{0.58} & \textbf{0.59} &   \textbf{0.62}   & \textbf{0.60}     & \textbf{0.64}    & \textbf{0.67}     \\
\hline
\end{tabular}
\end{table*}

\section{Ablation Study}
\subsection{Effect of Prompt Optimization}
\begin{figure}[htbp]
    \centering
    \includegraphics[width=\linewidth]{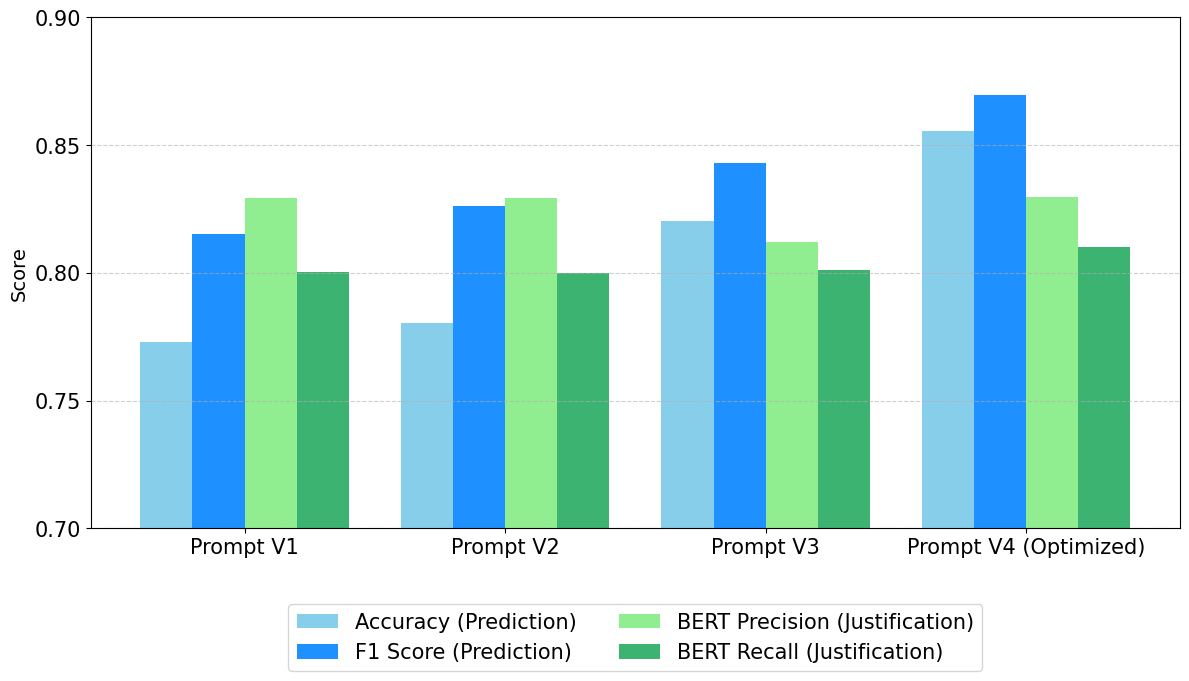}
    \caption{
        Performance of prompt variants on classification and justification tasks. 
        Optimized Prompt V4 achieves the best prediction scores, while justification quality remains stable, highlighting the impact of prompt optimization on multitask performance.
    }
    \label{fig:prompt_ablation}
\end{figure}

We systematically evaluated four prompt variants ranging from basic instructions (Prompt V1) to a fully optimized format (Prompt V4), comparing their impact on both prediction and justification performance. A clear trend emerges in Figure \ref{fig:prompt_ablation}: as the prompt design improves, models yield higher performance across the board. The optimized prompt (Prompt V4) delivers the best F1 score and accuracy on the classification task, outperforming the initial prompt (Prompt V1) by over 5\% absolute F1 improvement.

Interestingly, while the justification performance remains relatively stable across prompts, Prompt V4 still retains the highest BERT precision (0.83), suggesting that the optimized structure enhances not only decision accuracy but also the clarity and correctness of explanations. This confirms that prompt optimization plays a critical role in maximizing multitask alignment: better-structured prompts lead to more faithful learning of both the task objective and the reasoning behind predictions.

\subsection{Effect of LoRA Rank}

\begin{figure}[h]
    \centering
    \includegraphics[width=0.85\linewidth]{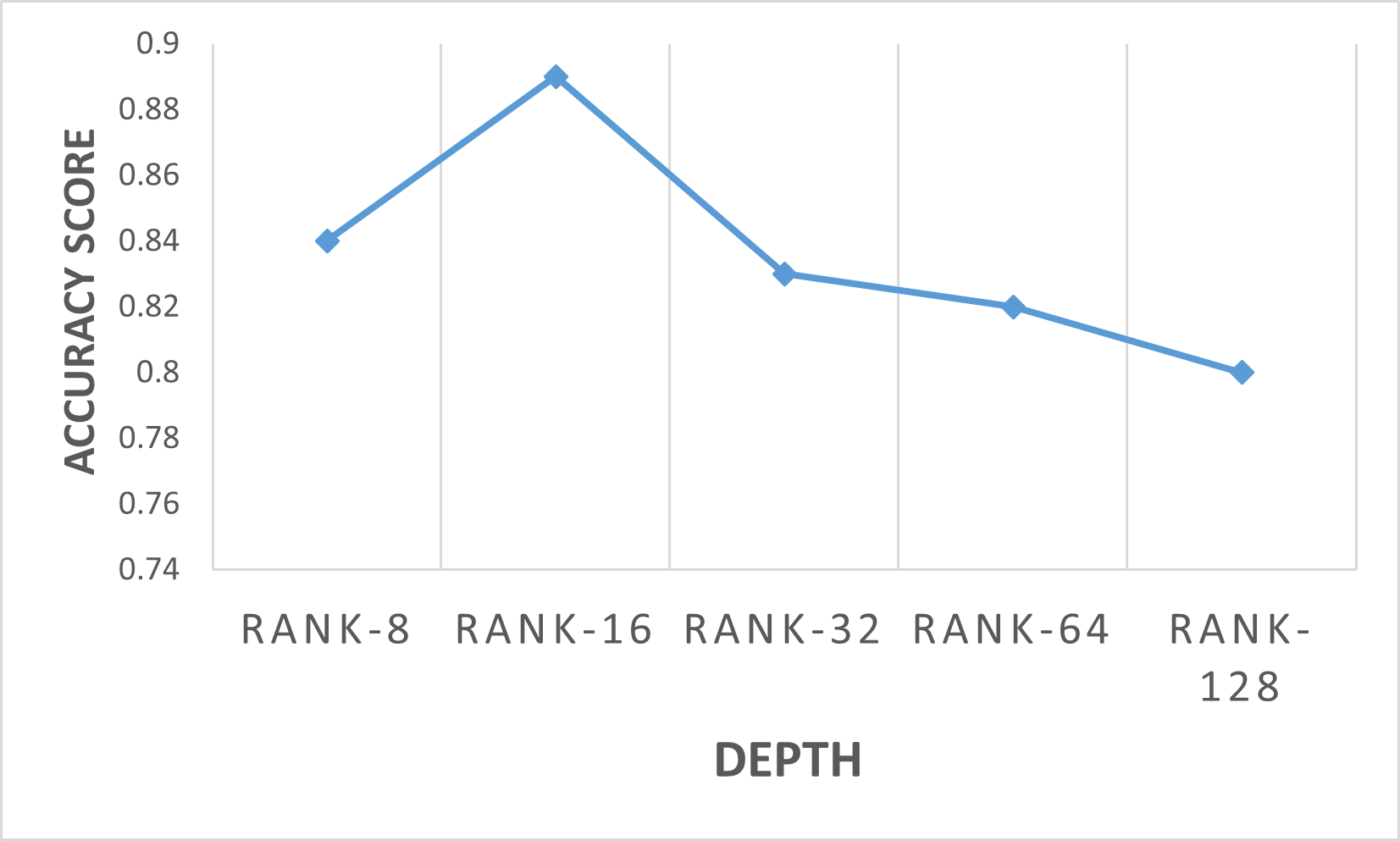}
    \caption{Effect of varying LoRA rank on accuracy. The best performance is achieved at rank-16, while both lower and higher ranks lead to reduced accuracy.}
    \label{fig:lora_rank}
\end{figure}

We further investigated the impact of varying the LoRA rank, which determines the adaptation capacity of low-rank adapters. As shown in Figure~\ref{fig:lora_rank}, the performance does not increase monotonically with rank size. The model achieves its best accuracy at rank-16, which is our default setup, reaching nearly 0.89, while both smaller (rank-8) and larger settings (rank-32, rank-64, rank-128) show reduced performance. This suggests that too little capacity (rank-8) limits expressiveness, whereas excessive capacity leads to overfitting or training instability on our dataset. Overall, the results highlight that moderate adapter size provides the best trade-off between efficiency and accuracy for structured business reasoning.

\section{Discussion, Implications, Limitations, and Future Work}
Our model provides data-driven explanations and our explanation module consistently highlights three signals as the strongest correlates of startup success: (i) \emph{funding levels}, (ii) \emph{number of investors} (syndication breadth), and (iii) \emph{executive team size}. In the following, we discuss why these factors plausibly matter and outline implications.

\paragraph{Funding levels}
Prior finance and entrepreneurship research links venture financing intensity to firm growth and exit potential. VC-backed firms scale faster (employment, sales) and exhibit lower early failure rates than non-VC firms matched, consistent with capital that allows commercialization and organizational expansion \citep{puri2012life}. At exit, venture backing and stronger pre-exit growth increase the likelihood of IPO relative to acquisition \citep{bayar2012valuationPremium,bayar2011ipoOrAcqTheory}. Classic certification work also shows that the involvement of reputable VC reduces information asymmetry in IPO \citep{megginson1991certification}. 
\emph{Implication:} the prominence of funding in our explanations is directionally consistent with theory and evidence: larger or sustained financing can proxy resource sufficiency and outside certification. While these patterns are robust, capital is not mechanistically causal; overfunding can amplify risk or misallocation in some regimes; therefore, decision makers should combine this signal with fundamentals (traction, margins, governance).

\paragraph{Number of investors (syndication breadth)}
Venture syndication is theorized to improve project selection (a second informed opinion) and/or add complementary value post-investment; empirically, broader syndicates and better networked VCs are associated with stronger results \citep{brander2002syndication,hochberg2007whom}. Using Crunchbase-scale data, collaborative experience within VC syndicates predicts exit paths \citep{pahnke2021amj}. 
\emph{Implication:} when the model highlights ``number of investors,'' it likely captures both (a) selection/certification and (b) access to broader networks, customers, and follow-up capital. For practitioners, the \emph{composition} of the syndicate (experience, reputation, network centrality), not just the count, is the lever worth auditing.

\paragraph{Executive team size}
A large literature links the resources of the founding / top management team (TMT) and professionalization to growth and survival. The attributes of the founding team are correlated with the growth of the company in the technology sectors \citep{eisenhardt1990organizational}; Meta-analytic evidence shows that human capital is positively (though modestly) related to entrepreneurial success \citep{unger2011human}. VC involvement accelerates professionalization (e.g., outside CEO hiring, role formalization), which in turn supports scaling and readiness for exit \citep{hellmann2002venture}. While the quality and role structure of teams are well studied, our finding that team size itself repeatedly surfaces as a top predictive feature on Crunchbase (controlling for funding, industry, and firm age) is a data-driven complement. It plausibly proxies managerial capacity and functional coverage, though causality can run both ways (funding $\rightarrow$ hiring $\rightarrow$ size).
\emph{Implication:} for founders, timely role specialization and measured team expansion are associated with higher exit odds; for investors, team footprint, together with the mix of experience, can inform selection and monitoring.

Implications for Management and Policy: 
For entrepreneurs, \emph{sequenced capitalization}, deliberate \emph{syndicate assembly}, and \emph{professionalized team build-out} are consistent with higher exit odds. For investors, the highlighted signals support triage and monitoring, especially when paired with calibration and bias checks. For policymakers, programs that expand capital access, mentor/investor networks, and managerial training can shift the distribution of scalable outcomes in regional ecosystems.

Our analysis faces several limitations. First, regarding causal inference: the identified factors are correlates, not necessarily causal levers, since funding and team size are jointly determined with performance. Second, with respect to data quality: Crunchbase coverage may vary by sector, geography, and vintage, which may affect the relative importance of features. Third, regarding scope: our outcome definition (IPO/acquisition) does not capture alternative success notions (e.g., sustainable profitability, private unicorn status). Finally, the faithfulness of the explanation is bounded by the model and features used.

In summary, our analysis reveals that funding levels, syndication breadth, and executive team size emerge as consistent predictors of startup success in the Crunchbase data. While these findings align with established theory, they also highlight the value of large-scale empirical analysis in quantifying the relative importance of different success factors. As the entrepreneurial ecosystem continues to evolve, such data-driven insights become increasingly valuable for founders, investors, and policymakers alike.

For \emph{future work}, we plan two main extensions. First, methodological improvements: we will evaluate across additional time periods and geographies to assess generalization, and incorporate richer signals (e.g., founder/investor networks, news, regulatory filings) with explicit temporal dynamics. Second, robustness enhancements: we will study calibration and fairness to mitigate bias, and integrate causal and counterfactual analyses for ``what-if'' reasoning.

\section{Conclusion}
We present CrunchLLM, a domain-adapted large language model framework to predict startup success from Crunchbase data. By jointly leveraging structured company attributes and unstructured textual narratives, and by employing parameter-efficient fine-tuning and prompt optimization, CrunchLLM substantially outperforms traditional machine learning baselines and untuned LLMs. Our experiments show a precision that exceeds 90\% in the IPO / acquisition prediction task, and the model reasoning traces provide interpretable justifications that support transparent decision-making for investors, researchers and policy makers.

\section*{Acknowledgements}
This research is supported in part by the NSF under Grant IIS 2327113 and ITE 2433190, the NIH under Grants R21AG070909 and P30AG072946, and the National Artificial Intelligence Research Resource (NAIRR) Pilot NSF OAC 240219, and Jetstream2, Bridges2, and Neocortex resources. We thank the University of Kentucky Center for Computational Sciences and Information Technology Services Research Computing for their support and use of the Lipscomb Compute Cluster and associated research computing resources. 

\bibliographystyle{IEEEtran}
\bibliography{sample}

\end{document}